%% file: emnlp2020.tex
\documentclass[11pt,a4paper]{article}
\usepackage[hyperref]{emnlp2020}

\usepackage{amsmath}
\usepackage{amssymb}
\usepackage{booktabs}
\usepackage{caption}
\usepackage{graphicx}
\usepackage{latexsym}
\usepackage{mathtools}
\usepackage{multirow}
\usepackage{subcaption}
\usepackage[super]{nth}
\usepackage{times}
\usepackage{xcolor}

\usepackage{microtype}

\aclfinalcopy %

\newcommand{\constraintCurveName}{constraint robustness curve }
\newcommand{\constraintCurveMetricFull}{Adjusted Constraint C-Statistic (ACCS) }
\newcommand{\constraintCurveMetric}{ACCS }

\title{Second-Order NLP Adversarial Examples}

\author{
	John X. Morris \\
	University of Virginia \\
	\texttt{jm8wx@virginia.edu} \\
}

\date{}

\begin{document}

\maketitle
\input{sections/00-Abstract}

\input{sections/01-Introduction}

\input{sections/02-Definitions}
\input{sections/03-Metrics}
\input{sections/04-Experiments}

\input{sections/05-Results}

\input{sections/05-01-Results-Tables-Figures}
\input{sections/06-Discussion}
\input{sections/07-Related-Work}
\input{sections/08-Conclusion}

\clearpage

\ifaclfinal
\section*{Acknowledgments}

This work arose out of a series of discussions with Eli Lifland about adversarial examples in NLP. Thanks to him and many others, including Jeffrey Yoo, Jack Lanchantin, Di Jin, Yanjun Qi, and Charles Frye, for engaging in similar discussions, which ranged from empirical to downright philosophical.
\fi

\bibliographystyle{acl_natbib}
\bibliography{emnlp2020}

\clearpage

\appendix
\input{sections/99-Appendix.tex}

\end{document}

%% file: sections/00-Abstract.tex
\begin{abstract}

Adversarial example generation methods in NLP rely on models like language models or sentence encoders to determine if potential adversarial examples are valid. In these methods, a valid adversarial example fools the model being attacked, and is determined to be semantically or syntactically valid by a second model. Research to date has counted all such examples as errors by the attacked model. We contend that these adversarial examples may not be flaws in the attacked model, but flaws in the model that determines validity. We term such invalid inputs second-order adversarial examples. We propose the constraint robustness curve, and associated metric ACCS, as tools for evaluating the robustness of a constraint to second-order adversarial examples. To generate this curve, we design an adversarial attack to run directly on the semantic similarity models. We test on two constraints, the Universal Sentence Encoder (USE) and BERTScore. Our findings indicate that such second-order examples exist, but are typically less common than first-order adversarial examples in state-of-the-art models. They also indicate that USE is effective as constraint on NLP adversarial examples, while BERTScore is nearly ineffectual. Code for running the experiments in this paper is available \href{https://github.com/jxmorris12/second-order-adversarial-examples}{here}.
\end{abstract}

%% file: sections/01-Introduction.tex
\section{Introduction}

If an imperceptible change to an input causes a model to make a misclassification, the perturbed input is known as an adversarial example \cite{Goodfellow2014-eq}. In domains with continuous inputs like audio and vision, whether such a change is considered ``imperceptible'' can be  easily measured: A change to an image may be considered imperceptible (and thus a valid adversarial example) if the resulting image is no more than some fixed distance away in pixel space \cite{AESurvey-Chakraborty2018-hz}. 

We refer to the function that determines imperceptibility as the constraint, $C$. For input $x$ and perturbation $x_{adv}$, if $C(x,x_{adv})$ is true, $x_{adv}$ is a valid perturbation for $x$.

Different domains call for different constraints. In vision, a common constraint is $\ell_{\infty}(x, x_{adv})$, the maximum pixel-wise distance between image $x$ and its perturbation $x_{adv}$ \cite{Goodfellow2014-eq}. In audio, a common constraint is $|dB(x) - dB(x_{adv})|$, the distortion in decibels between audio input $x$ and perturbation $x_{adv}$ \cite{Audio-Carlini2018-zy}. Both constraints are easily computed, well-understood, and correlate with human perceptual distance.

\begin{figure}
\centering
\begin{subfigure}{.9\columnwidth}
  \includegraphics[width=\linewidth]{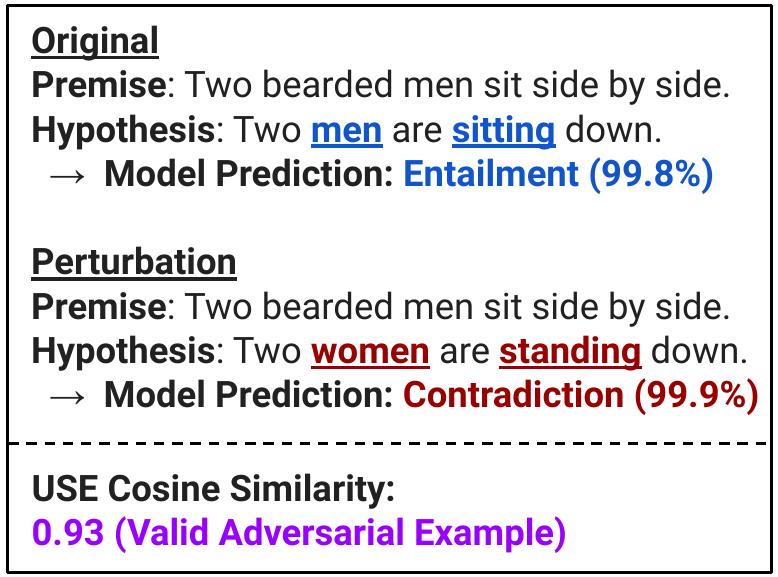}
\end{subfigure}

\caption{A second-order adversarial example in NLP. Although the perturbation has different meaning than the original (and the entailment model correctly predicts a contradiction), the sentence encoding similarity does not reflect this change. Current NLP adversarial example generation methods would incorrectly consider this a flaw in the entailment model.}

\label{fig:intro-use-ae-example}

\end{figure}

Choosing the correct constraint is not always so straightforward. In discrete domains like language, there is no obvious choice. In fact, the field lacks consensus on even the meaning of ``imperceptibility''. Different adversarial attacks have used different definitions of imperceptibility \cite{Zhang2020-NLPAE-Survey-cs}. One common definition \cite{Alzantot2018-ti,TextFooler-Jin2019-re,PWWS-Ren2019-hs,BAE-Garg2020-jq} is imperceptibility with respect to meaning: $C(x,x_{adv})$ is true if $x_{adv}$ retains the semantics of $x$.

With this definition, a perturbation $x_{adv}$ is determined to be a valid adversarial example if it simultaneously fools the model and retains the semantics of $x$. This formulation is problematic because measuring semantic similarity is an open problem in NLP. As a consequence, many adversarial attacks use a \textit{second NLP model} as a constraint, to determine whether or not $x_{adv}$ preserves the semantics of $x$.

Just like the model under attack, the semantic similarity model is vulnerable to adversarial examples. So when this type of attack finds a valid adversarial example, it is unclear which model has made a mistake: was it the model being attacked, or the model used to enforce the constraint? 

In other words, it is possible that the semantic similarity model improperly classified $x_{adv}$ as preserving the semantics of $x$. We refer to these flaws in constraints as \textbf{second-order adversarial examples}. Figure \ref{fig:intro-use-ae-example} shows a sample second-order adversarial example. Second-order adversarial examples have been largely ignored in the literature on NLP adversarial examples to date. 

Now that we are aware of the existence of second-order adversarial examples, we seek to minimize their impact. How can we measure a given constraint's susceptibility to second-order adversarial examples? We suggest one such measurement tool: the \textbf{constraint robustness curve} and its associated metric ACCS.

We then develop an adversarial example generation technique for finding examples that fool these semantic similarity models. Our findings indicate that adversarial examples for these types of models exist, but are less likely than adversarial examples that fool other NLP models. 

Along the way, we compare the Universal Sentence Encoder (USE) \cite{USE-Cer2018-ku}, a sentence encoder commonly used as a constraint for NLP adversarial examples, with BERTScore \cite{BERTScore-Zhang2019-no}, a metric that outperforms sentence encoders for evaluating text generation systems.

The main contributions of this work can be summarized as follows:
\begin{enumerate}
	\item We formally define second-order adversarial examples, a previously unaddressed issue with the problem statement for semantics-preserving adversarial example generation in NLP.
	\item We propose Adjusted Constraint C-Statistic (ACCS), the normalized area under the constraint robustness curve, as a measurement of the efficacy of a given model as a constraint on adversarial examples. 
	\item We run NLP adversarial attacks not on models fine-tuned for downstream tasks, but on semantic similarity models used to regulate the adversarial attack process.  We show that they are [robust|not robust]. Across the board, USE achieves a much higher ACCS, indicating that USE is a more robust choice than BERTScore for constraining NLP adversarial perturbations.
\end{enumerate}

%% file: sections/02-Definitions.tex
\section{Second-order adversarial examples}

To create natural language adversarial examples that preserve semantics, past work has implemented the constraint using a model that measures semantic similarity \cite{BAE-Garg2020-jq,Alzantot2018-ti,TextBugger-Li2018-vi,TextFooler-Jin2019-re}. For semantic similarity model $S$, original input $x$, and adversarial perturbation $x_{adv}$, constraint $C$ can be defined as defined as: 
\begin{equation}
	C(x, x_{adv}) := S(x, x_{adv}) \geq \epsilon
	\end{equation}

where $\epsilon$ is a threshold that determines semantic similarity. If their semantic distance is higher than some threshold, the perturbation is considered a valid adversarial example. 

Using such a constraint in an untargeted attack on classification model $F$, the attack goal $G$ function can be written as:

\begin{flalign}
	G(x, x_{adv}) &:= (F(x) \neq F(x_{adv})) \wedge C(x, x_{adv}) \nonumber  \\
  	&:= (F(x) \neq F(x_{adv})) \wedge (S(x, x_{adv} \geq \epsilon)  
\end{flalign}

Here, $x_{adv}$ is a valid adversarial example when both criteria of the goal are fulfilled: $F$ produces a different class output for $x_{adv}$ than for $x$, and $C(x, x_{adv})$ is true. This type of joint goal function is common in NLP adversarial attacks \cite{Zhang2020-NLPAE-Survey-cs}. 

It is possible that these constraints evaluate the semantic similarity of the original and perturbed text incorrectly. If the semantic similarity score is too low, then $x_{adv}$ will be rejected by the algorithm; if the score is too high, then the algorithm will consider $x_{adv}$ a valid adversarial example. 

If $S(x,x_{adv})$ is too high, $x_{adv}$ is incorrectly considered a valid adversarial example: a flaw in model $F$. However, since semantics is not preserved from $x$ to $x_{adv}$, there is no reason to assume that $F(x)$ should be consistent with $F(x_{adv})$. The flaw is actually in $S$, the semantic similarity model that erroneously considered $x_{adv}$ to be a valid adversarial example.

For adversarial attacks on model $F$ using a constraint determined by model $S$, we suggest the following terminology:
\begin{itemize}
	\item \textbf{First-order adversarial examples} are perturbations that are correctly classified as imperceptible by $S$, and fool $F$.
	\item \textbf{Second-order adversarial examples} fool $S$, the model used as a constraint. Regardless of the output of $F$, these are adversarial examples for $S$. 
\end{itemize}

In the next section, we suggest a method for determining the vulnerability of $S$ to second-order adversarial examples.

%% file: sections/03-Metrics.tex
\section{Constraint robustness curves and \constraintCurveMetric}

In this section, we propose the constraint robustness curve, a method for analyzing the robustness, or susceptibility to second-order adversarial examples, of a given constraint.

\label{s3:measuring-robustness-curves}

\begin{figure}

\begin{subfigure}{\linewidth}
  \includegraphics[width=\linewidth]{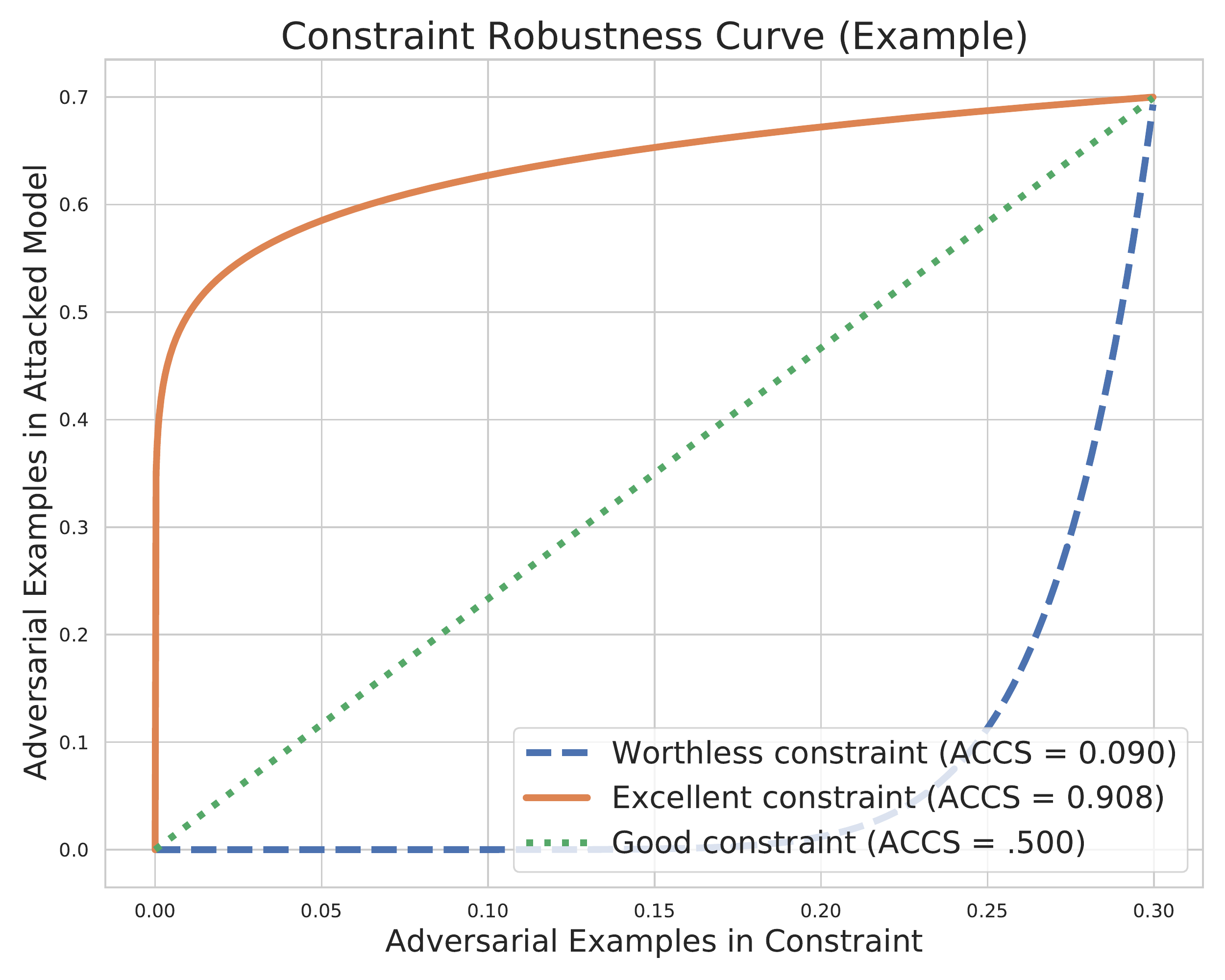}
\end{subfigure}

\caption{An example constraint robustness curve. \constraintCurveMetric is defined as the normalized area under the constraint robustness curve.}

\label{s3:\constraintCurveMetric-example}

\end{figure}

Each semantic similarity model may produce scores on a different scale, varying  the best $\epsilon$ for preservation of semantics. As such, we cannot fairly compare two models at the same values of $\epsilon$. 

However, the problem of comparing two binary classifiers that may have different threshold scales is common in machine learning \cite{ROC-Hajian-Tilaki2013-oa}. Inspired by the receiver operating characteristic (ROC) curve for binary classifiers, we propose the \textbf{constraint robustness curve}, a plot of first-order vs. second-order adversarial examples as constraint sensitivity varies. To create the \constraintCurveName for semantic similarity model $S$ and threshold $\epsilon$, we plot the number of true positives (first-order adversarial examples, found using $S$ as a constraint) vs. false positives (second-order adversarial examples, found by attacking $S$ directly).

The \constraintCurveName can be interpreted similarly to an ROC curve. An effective constraint will allow many true positives (first-order adversarial examples) before many false positives (second-order adversarial examples). The model that produces a curve with a higher AUC (area under the constraint robustness curve) is better at distinguishing valid from invalid adversarial examples, and less susceptible to second-order adversarial examples.

When $\epsilon=0$, $C(x, x_{adv})$ is always true. But even when the constraint accepts all possible $x_{adv}$, some attacks may still fail. So unlike a typical ROC curve, which is bounded between $0$ and $1$ on both axes, the \constraintCurveName is bounded on each axis between $0$ and the maximum attack success rate (when $\epsilon=0$). We suggest normalizing to bound the score between 0 and 1. 

We call the resulting metric \textbf{\constraintCurveMetricFull}\footnote{\textit{C-statistic} is another name for AUC.}. \constraintCurveMetric is defined as the area under the constraint robustness curve normalized by the maximum first- and second-order success rate. Figure \ref{s3:\constraintCurveMetric-example} shows an example of a \constraintCurveName for a toy problem. (The area under the green dashed curve is $0.105$; after normalizing by the maximum first- and second-order attack success rates of $0.7$ and $0.3$, we find $ACCS = 0.5$.)

There is one crucial difference between interpreting an ROC curve and a constraint robustness curve. A naive binary classifier will guess randomly and achieve as many false positives as true positives, and an AUC of $0.5$. A naive constraint will yield all second-order adversarial examples at the same threshold, and garner an ACCS of $0.0$.

To create such a curve, we must devise methods for generating both first-order and second-order adversarial examples. In the following section, we propose an attack for each purpose.

%% file: sections/04-Experiments.tex
\section{Generating first and second-order adversarial examples}

To calculate $ACCS(S, \epsilon)$ for each $S$ and $\epsilon$, we design two attacks: one to calculate the number of first-order adversarial examples, and one to calculate the number of second-order adversarial examples. In Section \ref{s5:results}, we run the attacks across a variety of models and datasets and examine their constraint robustness curves.

\subsection{Generating first-order adversarial examples}
\label{s3:generating-first-order}

To measure the number of first-order adversarial examples allotted by a semantic similarity model for a given value of $\epsilon$, we can run any standard adversarial attack that uses the semantic similarity model as a constraint.

We devise a simple attack to generate adversarial examples for some classifier $F$. We choose untargeted classification, the goal of changing the classifier's output to any but the ground-truth output class, as the goal function. To generate perturbations, we swap words in $x$ with their synonyms from WordNet \cite{WordNet-Miller1995-ft}.

Simply swapping words with synonyms from a thesaurus would frequently create ungrammatical perturbations (even though they may be semantically similar to the originals). To better preserve grammaticality, we enforce an additional constraint, requiring that the log-probability of any replaced word not decrease by more than some fixed amount, as according to the GPT-2 language model \cite{GPT2-Radford2019-kb}. (This is similar the language model perplexity constraints used in the NLP attacks of \citet{Alzantot2018-ti} and \cite{Kuleshov2018-NLPAEs-di}.)

As an additional constraint, the attack filters potential perturbations using the semantic similarity model to ensure that $S(x, x_{adv}) \geq \epsilon$.

Finally, we choose greedy with word importance ranking as our search method \cite{DeepWordBug-Gao2018-gt}. We can use these four components (goal function, transformation, constraints, and search method) to construct an adversarial attack to generate adversarial examples for any NLP classifier \cite{TextAttack-Morris2020-qp}.

\subsection{Generating second-order adversarial examples}
\label{s3:generating-second-order}

Generating adversarial examples for classification model $F$ is a well-studied problem. But how do we generate perturbations that fool $S$, a semantic similarity model?

We first note what these adversarial examples might look like. Our goal is to find `false positives' where a semantic similarity model incorrectly indicates that semantics is preserved. Specifically, we want to find some $(x, x_{adv})$ where $S(x, x_{adv}) \geq \epsilon$, even though we know $x_{adv}$ does not preserve the semantics of $x$.

To generate such perturbations, we design a transformation with the goal of changing the meaning of an $x$ as much as possible (instead of preserving its meaning). At each step of the adversarial attack, instead of replacing words with their synonyms, we replace words with their antonyms, also sourced from WordNet \cite{WordNet-Miller1995-ft}.

Next, we need to establish a goal function that perturbations must meet to be considered adversarial examples for a given semantic similarity metric. We establish the following goal function:

\begin{equation}
	\begin{split}
	G(x, x_{adv}) := (S(x, x_{adv}) \geq \epsilon) \ \wedge \\
		((\sum_{i} x[i] \neq x_{adv}[i]) \geq \gamma) 
	\end{split}
\end{equation}

Here, $x[i]$ represents the $i\textsuperscript{th}$ word in sequence $x$, and $\gamma$ represents the minimum number of words that must be changed for the attack to succeed. 

With our goal function, perturbation $x_{adv}$ is a valid adversarial example if it differs by at least $\gamma$ words from $x$, but its semantic similarity to $x$ is still higher than $\epsilon$. If $\gamma$ words are substituted with antonyms, as $\gamma$ increases, we can say with high certainty that semantics is not preserved. In this case, the semantic similarity model \textit{should} produce a value smaller than $\epsilon$.

As in \ref{s3:generating-first-order}, we apply a second constraint, using GPT-2 to ensure antonyms substituted are likely in their context. For the search method, we use beam search, as it does a better job finding adversarial examples when the set of valid perturbations is sparse \cite{HotFlip-Ebrahimi2017-ca}. 

A sample output of this attack (where $\gamma=2$) is shown in Figure \ref{fig:intro-use-ae-example}.

%% file: sections/05-Results.tex
\section{Experiments}
\label{s5:results}

\begin{figure}

\begin{subfigure}{.48\textwidth}
  \includegraphics[width=\linewidth]{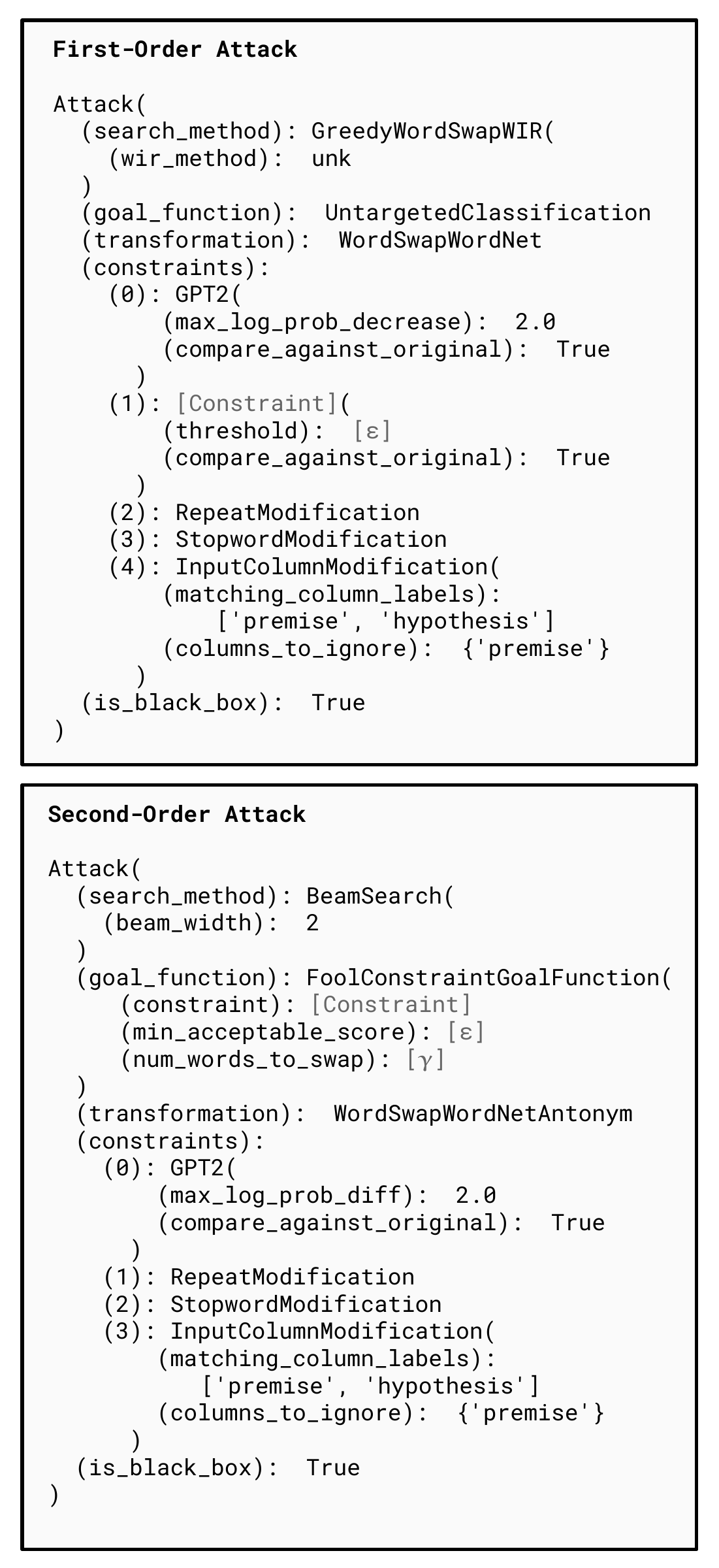}
\end{subfigure}

\caption{Attack prototypes generated for attacks run in TextAttack. The top shows the first-order attack, run against a classification model using the semantic similarity model as a constraint. The bottom shows the second-order attack, run directly against a semantic similarity model. During experiments, {\color{darkgray}[Constraint]} is either USE or BERTScore, {\color{darkgray}[$\epsilon$]} is varied from $0.5$ to $1$ or $0.75$ to $1$, and {\color{darkgray}[$\gamma$]} is set to $3$.}

\label{s5:attack-prototypes}

\end{figure}

\subsection{Attack Prototypes}

We implemented our adversarial attacks using the TextAttack adversarial attack framework \cite{TextAttack-Morris2020-qp}. Figure \ref{s5:attack-prototypes} shows the attack prototypes of each attack, as constructed in TextAttack. 

As noted in the previous section, each attack used the GPT-2 language model to preserve grammaticality during word replacements; we disallowed word replacements that decreased in log-probability from the original word 2.0 or more. The other constraints in the attack prototype disallow multiple modifications of the same word, stopword substitutions, and, in the case of entailment datasets, edits to the premise. \footnote{It is standard for NLP attacks on entailment models to only edit the hypothesis \cite{Alzantot2018-ti,GAN-NLPAEs-Zhao2017-SameerSingh-zs, TextFooler-Jin2019-re}).}

\subsection{Semantic similarity models}
We tested two semantic similarity models as $S$:
\begin{itemize}
	\item The Universal Sentence Encoder (USE) \cite{USE-Cer2018-ku}, a model trained to encode sentences into fixed-length vectors. Semantic similarity between $x$ and $x_{adv}$ is measured as the cosine similarity of their encodings. This is consistent with NLP attack literature \cite{TextBugger-Li2018-vi,TextFooler-Jin2019-re,BAE-Garg2020-jq}.
	\item BERTScore \cite{BERTScore-Zhang2019-no}, an automatic evaluation metric for text generation. BERTScore computes a similarity score for each token in the candidate sentence with each token in the reference sentence using the contextual embedding of each token. According to human studies, BERTScore correlates better than other metrics (including sentence encodings) for evaluating machine translations. It also outperforms sentence encodings on PAWS \cite{PAWS-Yang2019-pj}, an adversarial paraphrase dataset where inputs have a similar format to NLP adversarial examples.
\end{itemize}

\subsection{Victim Classifiers}
To create constraint robustness curves, we ran each attack (first and second-order) while varying $\epsilon$ from $0.75$ to $1.0$ in increments of $0.01$. For the SST-2 dataset, which has some very short examples, we varied $\epsilon$ from $0.5$ to $1.0$ in increments of $0.02$. For the second-order attack, we fixed $\gamma = 3$. For our tests, we chose the following three datasets:

\begin{itemize}
	\item The Stanford Natural Language Inference (SNLI) Corpus, which contains labeled sentence pairs for textual entailment \cite{SNLI-Bowman2015-wc};
	\item The Stanford Sentiment Treebank v2 (SST-2) Corpus \cite{SST2-Socher2013-xf}, a phrase-level sentiment classification dataset;
	\item Rotten Tomatoes dataset \footnote{The Rotten Tomatoes dataset is sometimes called Movie Review, or MR, dataset.}, a sentence-level sentiment classification dataset \cite{RottenTomatoes-Pang+Lee:05a}.
\end{itemize}

For the first-order attack, we chose three target models fine-tuned on each dataset (total of nine models): BERT \cite{BERT-Devlin2018-wu}, ALBERT \cite{ALBERT-Lan2019-wo}, and DistilBERT \cite{DistilBERT-Sanh2019-vf}. All models used were pre-trained models provided by TextAttack \cite{TextAttack-Morris2020-qp}. More details about experimental setup are provided in \ref{app:experimental-details}.

\subsection{Results}

We sampled 100 examples from each test set of dataset for each attack. We repeated each attack twice, once using BERTScore and once using the Universal Sentence Encoder. In total, we ran 300 attacks.

Table \ref{table:accs-results} shows results for each model and dataset. Figure \ref{fig:attack-roc} shows the constraint robustness curve for each scenario.

Surprisingly, the Universal Sentence Encoder achieved a higher ACCS than BERTScore across all nine scenarios. This appears contradictory to the claims of \citet{BERTScore-Zhang2019-no} that ``BERTScore is more robust to challenging examples when compared to existing metrics''.

Additionally, at any given point, first-order adversarial examples are found over twice as often as second-order adversarial examples. This indicates that most adversarial examples found in NLP attacks may be first-order. This corroborates human studies from \citetext{Reevaluating-Morris2020-mb}, which showed that humans rate adversarial examples from the attacks of \citetext{Alzantot2018-ti} and \citetext{TextFooler-Jin2019-re} to preserve semantics around 65\% of the time.

\begin{figure}

\begin{subfigure}{.48\textwidth}
  \includegraphics[width=\linewidth]{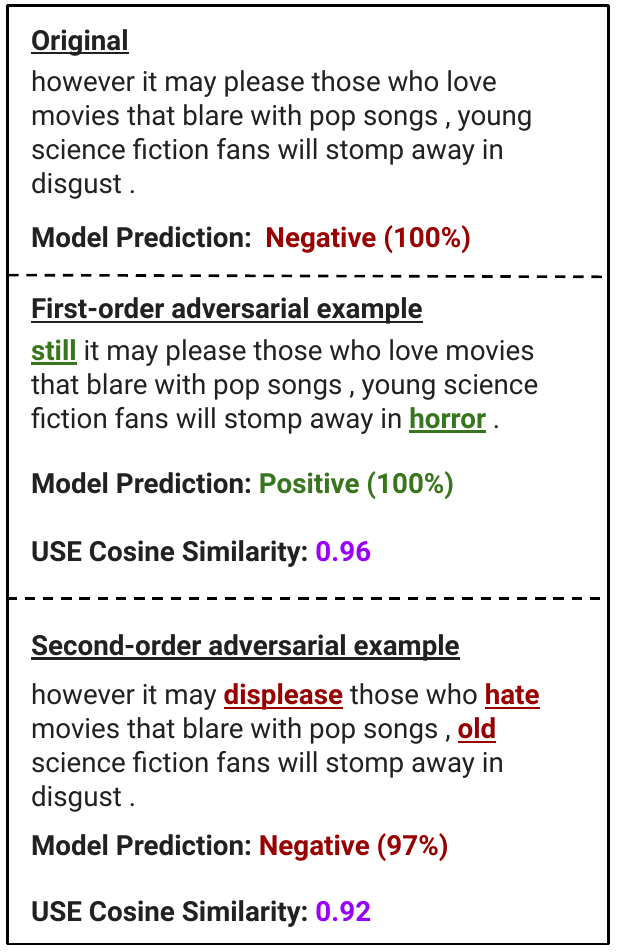}
\end{subfigure}

\caption{First-order and second-order adversarial examples generated by our attacks on BERT-base fine-tuned on the SST-2 dataset.}

\label{s5:attack-prototypes}

\end{figure}

%% file: sections/05-01-Results-Tables-Figures.tex
\begin{table}[]
\resizebox{\linewidth}{!}{%
\begin{tabular}{|l|l|l|l|}
\toprule
 &  & \multicolumn{2}{c|}{Constraint Under Attack} \\
Dataset & Target Model & BERTScore & USE \\
\midrule
\multirow{3}{*}{SNLI} & BERT & 0.590 & 0.730 \\
 & ALBERT & 0.569 & 0.678  \\
 & DistilBERT & 0.575 & 0.721 \\
 \hline
\multirow{3}{*}{SST-2} & BERT & 0.290 & 0.423  \\
 & ALBERT & 0.407 & 0.464  \\
 & DistilBERT & 0.405 & 0.511 \\
\hline
\multirow{3}{*}{Rotten Tomatoes (MR)} & BERT & 0.382 & 0.383 \\
 & ALBERT & 0.388 & 0.427 \\
 & DistilBERT & 0.448 & 0.466 \\
\bottomrule
\end{tabular}%
}

\caption{Results of first-order and second-order attacks on BERTScore and the Universal Sentence Encoder (USE). Values are ACCS, a measure of constraint robustness. A higher ACCS score indicates a better constraint. Across models and datasets, USE achieves a higher ACCS than BERTScore.}
\label{table:accs-results}
\end{table}

\begin{figure*}[ht]
\begin{subfigure}{\textwidth}
  \centering
  \includegraphics[width=\linewidth]{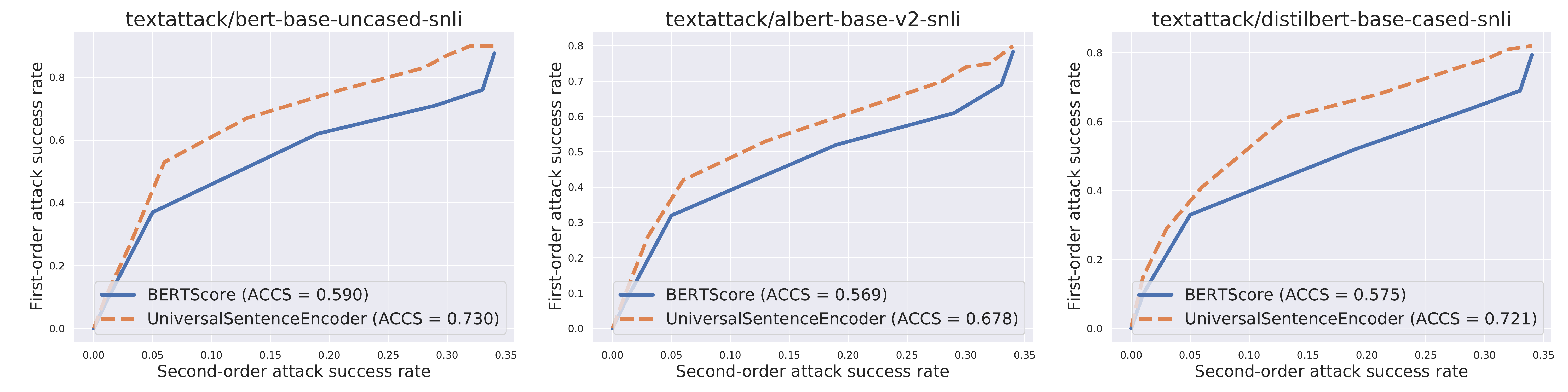}
  \label{fig:attack-roc-snli}
\end{subfigure}
\begin{subfigure}{\textwidth}
  \centering
  \includegraphics[width=\linewidth]{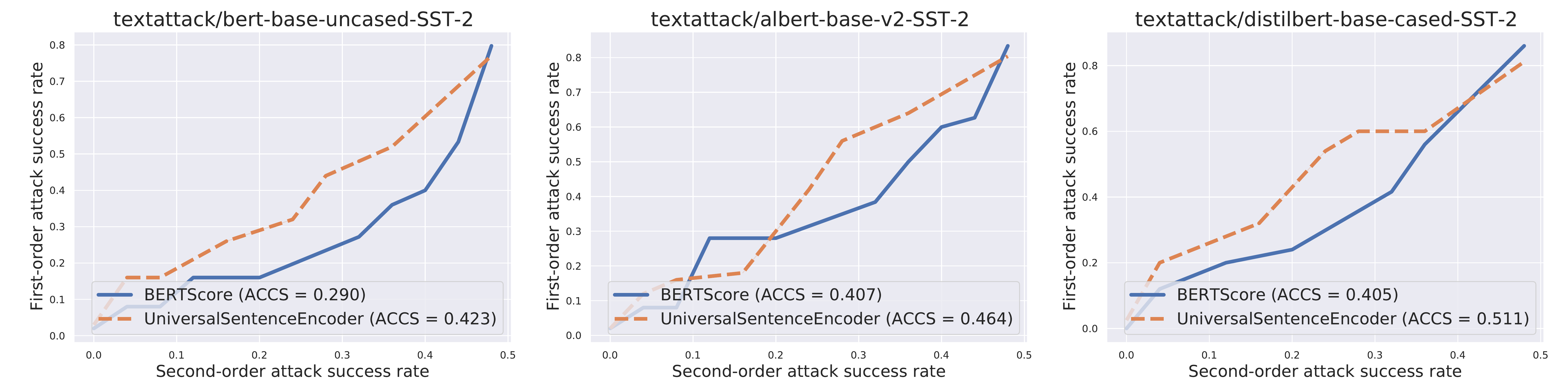}
  \label{fig:attack-roc-sst2}
\end{subfigure}
\begin{subfigure}{\textwidth}
  \centering
  \includegraphics[width=\linewidth]{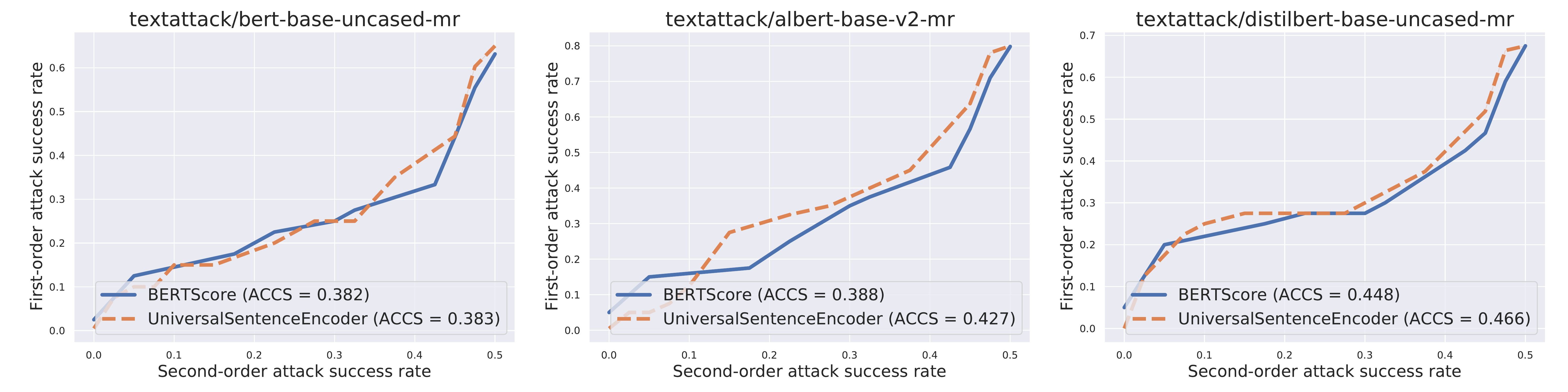}  
  \label{fig:attack-roc-rottentomatoes}
\end{subfigure}
\caption{Constraint robustness curves across attacks. The Universal Sentence Encoder finds more adversarial examples in each model while yielding fewer adversarial examples via second-order attacks. ACCS results are detailed in Table \ref{table:accs-results}.}
\label{fig:attack-roc}
\end{figure*}

%% file: sections/06-Discussion.tex
\section{Discussion}

\paragraph{Sentence length, $S$, and $\epsilon$.} As input $x$ grows in length, the a single word swap will have an increasingly smaller impact on $S(x, x_{adv})$. Some NLP attacks that use sentence encoders as a constraint have combatted this problem by measuring the sentence encodings within a fixed-length window of words around each substitution. For example, \citet{TextFooler-Jin2019-re} considers a window of 15 words around each substitution. We chose instead to encode the entire input, as both the Universal Sentence Encoder and BERTScore were trained using full inputs. %

\paragraph{Applications beyond NLP.} Table \ref{table:domain-metrics} lists examples of validity metrics across domains. To the best of our knowledge, no domains outside of NLP have suggested to use a deep learning model as a constraint \cite{AESurvey-Chakraborty2018-hz}. If adversarial attacks in other domains do decide to use deep learning models to measure imperceptibility, they can follow our method to compare imperceptibility models and evaluate their robustness.

\paragraph{The Catch-22 of second-order adversarial examples.} Any adversarial generation method for that employs an auxiliary model as a constraint may generate second-order adversarial examples. Although NLP is the only domain to use a model as a constraint thus far, this problem is likely to appear in other domains in the future. This makes the problem of second-order adversarial detection more important.

\paragraph{Towards better constraints on NLP adversarial examples.} Neither USE nor BERTScore scored especially high \constraintCurveMetric scores on any of the studied tasks. We leave it to future work to explore more choices of semantic similarity model and find one that is more suitable as a constraint on NLP adversarial examples.

%% file: sections/07-Related-Work.tex
\section{Related Work}

We can categorize adversarial attacks in NLP based on their chosen definition of imperceptibility: generally adversarial attacks in NLP aim either for \textit{visual} imperceptibility or in \textit{semantic} imperceptibility.

\paragraph{Visual imperceptibility.} These adversarial example generation techniques focus on character-level modifications that a fast-reading human may not notice. HotFlip \cite{HotFlip-Ebrahimi2017-ca} uses the gradient of a character-level classifier to guide the attack, and can often change the classifier output with a single flip. (HotFlip also studies word-level replacements, but only briefly.) Other works \cite{Belinkov-Synthetic-Natural2017-qq,DeepWordBug-Gao2018-gt,Pruthi2019-mx,JiaTypoRobustEncodings-Jones2020-tv} craft adversarial examples by inducing `typos' in the input sequence $x$, for example, by swapping two characters with one another, or shuffling the characters in an input. In these cases, imperceptibility is generally modeled using string edit distance, so second-order adversarial examples do not exist.

\paragraph{Semantic imperceptibility.} This work focuses on this class of NLP adversarial examples, in which $x_{adv}$ must preserve the semantics of $x$. Most work generates these $x_{adv}$ by swapping iteratively swapping words in $x$ with synonyms, and filtering by some model-based constraint \cite{Kuleshov2018-NLPAEs-di,PWWS-Ren2019-hs,TextFooler-Jin2019-re,BAE-Garg2020-jq}. Some alternative algorithms have been proposed: \citet{GAN-NLPAEs-Zhao2017-SameerSingh-zs} encode $x$ into a latent representation using a generative adversarial network, apply the perturbation to the latent vector, and decode to obtain $x_{adv}$. \citet{SEARS-Ribeiro2018-ue} craft `adversarial rules' (mappings from $x \rightarrow x_{adv}$) by a combination of back-translation and human evaluation. TextBugger \cite{TextBugger-Li2018-vi} crafts adversarial examples using word-level substitutions, but uniquely chooses between character-level perturbations (exploiting imperceptibility in appearance) and word-level synonym swaps (exploiting imperceptibility in meaning).

Although there have been many adversarial attacks proposed on NLP models \cite{Zhang2020-NLPAE-Survey-cs}, surprisingly few constraints have been explored. \citet{Alzantot2018-ti} was the first to propose the use of a language model as a constraint on grammaticality. \citet{Kuleshov2018-NLPAEs-di} uses both a language model to enforce grammaticality and skip-thought vectors \cite{SkipThought-Kiros2015-uc}, a form of sentence encoding, to enforce semantic preservation. Several attacks have used the Universal Sentence Encoder to enforce semantic preservation \cite{TextBugger-Li2018-vi,TextFooler-Jin2019-re,BAE-Garg2020-jq}.

\citet{Reevaluating-Morris2020-mb} categorized constraints on NLP adversarial examples into four groups: semantics, grammaticality, overlap, and non-suspicion. They also explored the effect of varying constraint threshold on the quality of generated adversarial examples, as judged by human annotators. \citet{Elephant-Xu2020-xi} examined the quality of generated adversarial examples based on different thresholds of attack success rate. However, neither study considered adversarial examples that may have arisen from constraints, or explored evaluation via running adversarial attacks on the constraints directly.

\begin{table}[]
	\resizebox{\linewidth}{!}{%
	\begin{tabular}{|l|l|}
	\toprule
	\textbf{Adversarial Example Domain} & \textbf{Constraint} \\
	\midrule
	Images \cite{Goodfellow2014-eq} & maximum $\ell_{\inf}$ norm \\
	\hline
	Audio \cite{Audio-Carlini2018-zy} & minimum distortion in Decibels (dB) \\
	\hline
	Graphs \cite{Graph-AEs-Wu2019-ha} & maximum number of edges modified \\
	\hline
	Text \cite{Survey-NLPAE-Zhang2020-cs} & minimum USE cosine similarity  \\
	\bottomrule
	\end{tabular}%
	}
	\caption{Examples of constraints across adversarial example domains. All metrics are calculated between the original input and any potentially valid adversarial perturbation.}
	\label{table:domain-metrics}
\end{table}

%% file: sections/08-Conclusion.tex
\section{Conclusion}
Work in generating adversarial examples in NLP has relied on outside models to evaluate imperceptibility. While useful, this inadvertently increases the size of the attack space. We propose methods for analyzing constraints' susceptibility to second-order adversarial examples, including the \constraintCurveMetric and associated \constraintCurveName metric. This requires us to design an attack specific to semantic similarity models. We demonstrate these methods with a comparison of two models used in constraints, the Universal Sentence Encoder and BERTScore. We would especially like to see future research examine constraint robustness curves across more constraints and different attack designs. We hope that future researchers can use our method when choosing constraints for NLP adversarial example generation.

%% file: sections/99-Appendix.tex
\section{Appendices}
\label{sec:appendix}

\subsection{Experimental Details}
\label{app:experimental-details}

\paragraph{Setup} All experiments were run using the TextAttack framework in Jupyter notebooks running in Google Colab using Tesla K80 GPUs. \footnote{Google Colab is a great resource, providing free, easy access to high-powered GPUs, but its timeout constraints can be frustrating and unpredictable. By the end of the project, this author shelled out \$9.99 for the high-octane \textit{Google Colab Pro}.}.

\paragraph{Models} The attacked models are pretrained models provided by TextAttack \cite{TextAttack-Morris2020-qp}. BERTScore and the Universal Sentence Encoder are also loaded through TextAttack. The pre-trained models are available on the \href{https://huggingface.co/models}{HuggingFace model hub} under the following names:

\begin{itemize}
	\item SNLI Dataset
		\begin{itemize}
			\item textattack/bert-base-uncased-snli
			\item textattack/albert-base-v2-snli
			\item textattack/distilbert-base-cased-snli
		\end{itemize}
	\item SST-2 Dataset
		\begin{itemize}
			\item textattack/bert-base-uncased-SST-2
			\item textattack/albert-base-v2-SST-2
			\item textattack/distilbert-base-cased-SST-2
		\end{itemize}
	\item Rotten Tomatoes Dataset
		\begin{itemize}
			\item textattack/bert-base-uncased-rotten-tomatoes
			\item textattack/albert-base-v2-rotten-tomatoes
			\item textattack/distilbert-base-uncased-rotten-tomatoes
		\end{itemize}
\end{itemize}

\subsection{Constraints tested on paraphrase datasets}

Before running adversarial attacks on USE and BERTScore, we compared their effectiveness on common paraphrase identification tasks.

USE and BERTScore each assign a semantic similarity score to each (original text, perturbed text) pair. A hard threshold determines whether a given score indicates a valid adversarial example. Above this threshold, the perturbed text is assumed to have preserved the semantics of the original input; below it, semantics is not preserved, and the perturbation is invalid. \citet{TextBugger-Li2018-vi} defines validity as a cosine similarity of $0.8$ or higher, as measured by USE. \citet{TextFooler-Jin2019-re} and \citet{BAE-Garg2020-jq} choose a lower USE threshold of $0.5$.

Current state-of-the-art attacks in NLP generate perturbations one word at a time: generally by swapping out a word with neighbors in the embedding space \cite{Alzantot2018-ti} or with synonyms provided by a thesaurus \cite{PWWS-Ren2019-hs}. Consequently, their adversarial perturbations share the lexical structure of the original inputs, with some words swapped out for synonyms. This implies that BERTScore would be a better fit for ensuring semantic preservation during these adversarial attacks, and less susceptible to second-order adversarial examples.

Our initial question was how USE and BERTScore compare on common datasets for paraphrase identification. When used as constraints on adversarial attacks, constraints that can more correctly distinguish paraphrases from non-paraphrases should be less vulnerable to second-order adversarial examples.

In the following subsections, we compare USE and BERTScore on two paraphrase datasets, QQP and PAWS, and then on Adversarial SNLI, on a custom dataset designed to resemble the format of NLP adversarial examples on the SNLI entailment dataset.

\subsubsection{Performance on paraphrase identification}

We evaluate USE and BERTScore on two common paraphrase datasets:

\begin{itemize}
	\item The \textbf{QQP (Quora Question Pairs)} dataset, which contains 400k real-world pairs of paraphrases and non-paraphrases collected during Quora question disambiguation.
	\item The \textbf{PAWS (Paraphrase Adversaries from Word Scrambling)} dataset, which contains 100k paraphrases and non-paraphrases. These examples originally come from the QQP paraphrases; non-paraphrases have been adversarially edited to change semantics while retaining high lexical overlap from the source. \cite{PAWS-Yang2019-pj}
\end{itemize}

We sampled 1000 examples from the QQP and PAWS test sets. All datasets are loaded using the \texttt{nlp} package from HuggingFace\footnote{See https://github.com/huggingface/nlp.}. The TextAttack library \cite{TextAttack-Morris2020-qp} is used to load pre-trained USE and BERTScore models and to run augmentation and adversarial attack experiments.

\begin{figure}

\begin{subfigure}{.5\textwidth}
  \includegraphics[width=\linewidth]{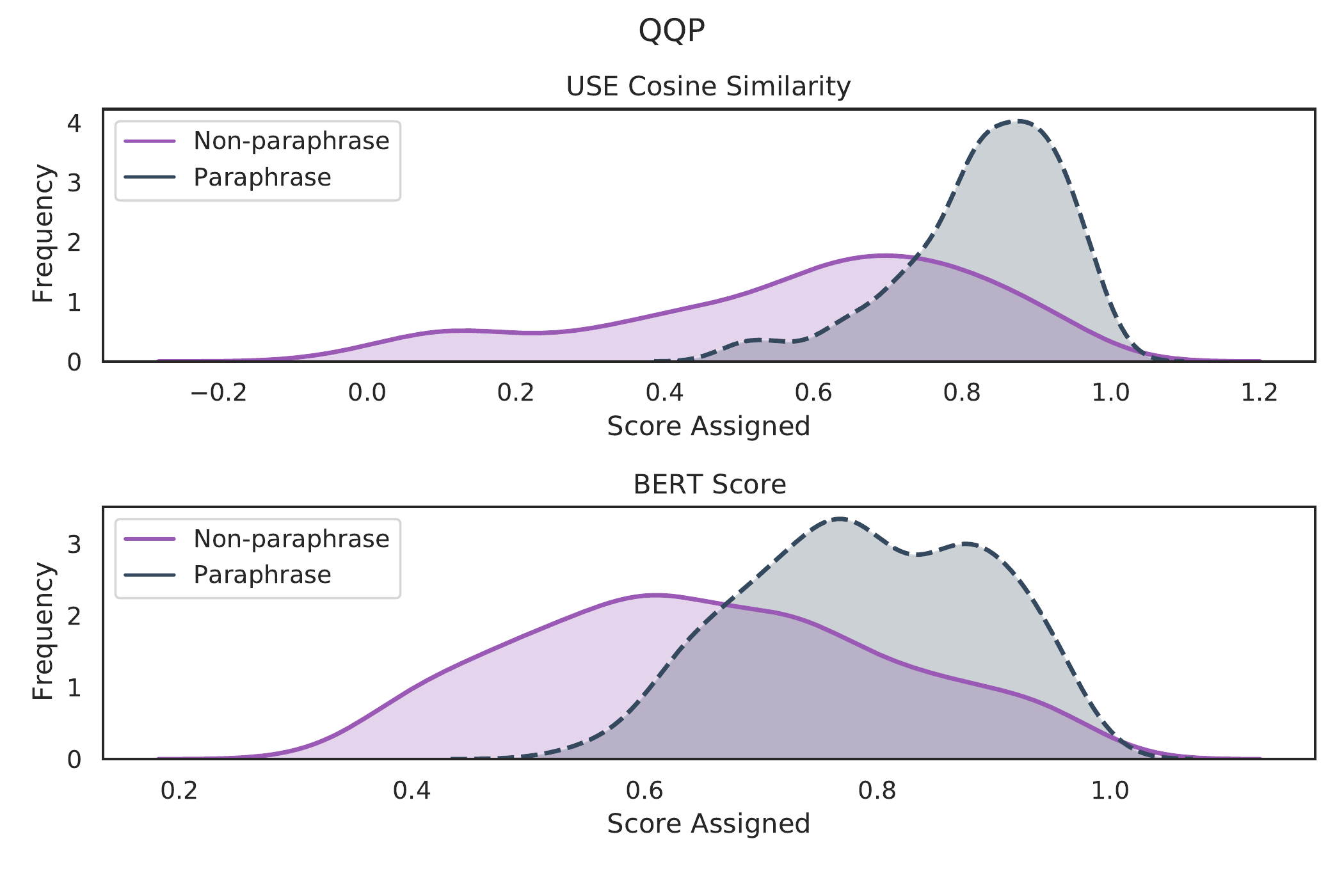}
\end{subfigure}

\begin{subfigure}{.5\textwidth}
  \includegraphics[width=\linewidth]{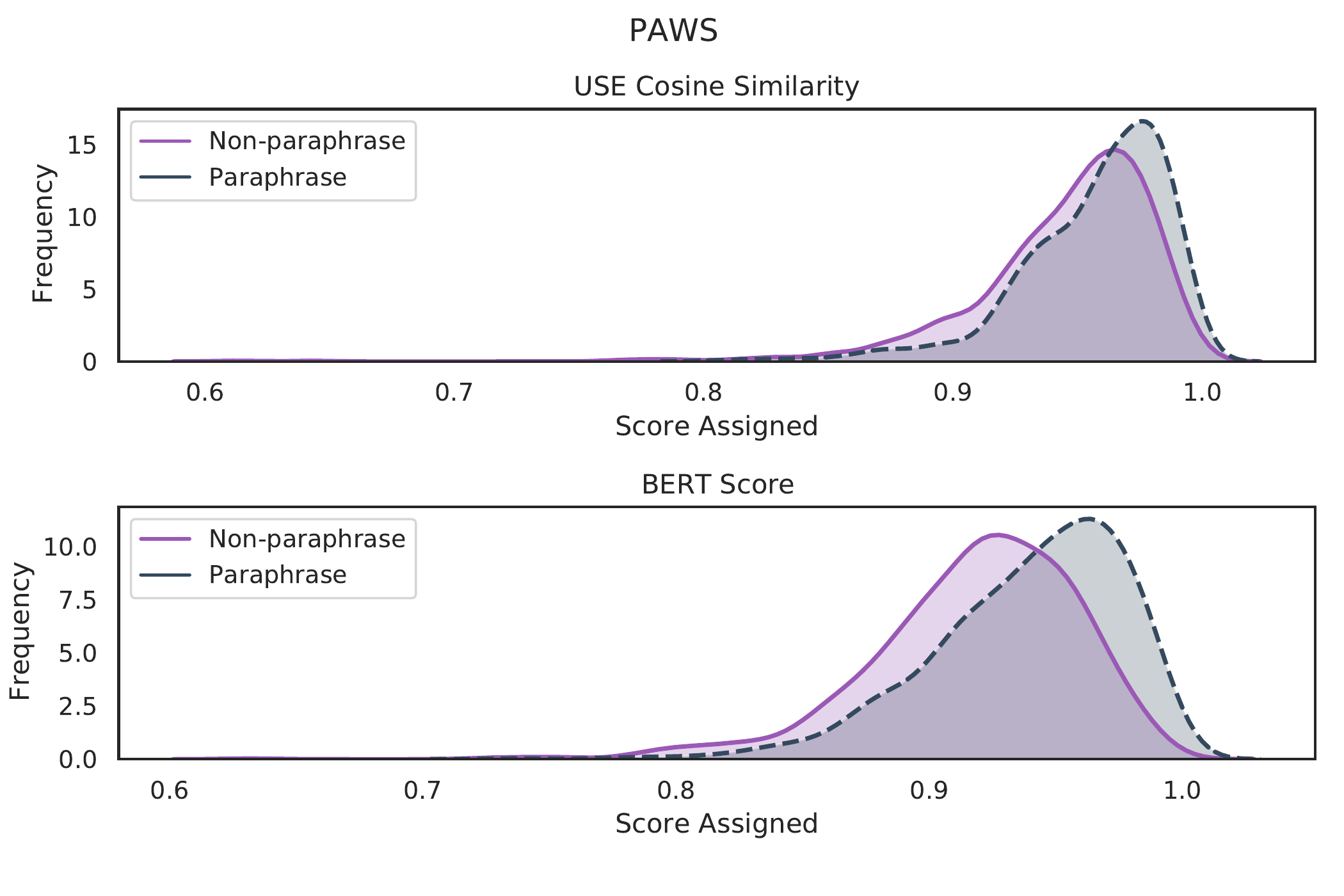}
\end{subfigure}

\caption{Distribution of scores assigned by BERTScore and the Universal Sentence Encoder (USE) on the QQP and PAWS datasets.}

\label{app:qqp-paws-distribution}

\end{figure}

Figure \ref{app:qqp-paws-distribution} shows the distributions of scores from each model (USE, BERTScore) on each dataset (QQP, PAWS). Both models exhibit some ability to distinguish paraphrases and non-paraphrases on QQP, but produce very similar scores for paraphrases and non-paraphrases on PAWS (with the non-paraphrases having slightly lower scores).

\begin{table}[]
	\begin{tabular}{lll}
		Dataset & USE & BERTScore \\
		QQP & 0.827 & 0.764 \\
		PAWS & 0.608 & 0.662 \\
		Adversarial SNLI & 0.635 & 0.710
	\end{tabular}
	\label{app:qqp-paws-adv-snli-auc-table}
	\caption{AUC Scores for BERTScore and the Universal Sentence Encoder on QQP, PAWS, and our Adversarial SNLI dataset. BERTScore shows an advantage PAWS and Adversarial SNLI, indicating that it is a more robust choice for constraining semantics during NLP adversarial example generation.}
\end{table}

\begin{figure}

\begin{subfigure}{.5\textwidth}
  \includegraphics[width=\linewidth]{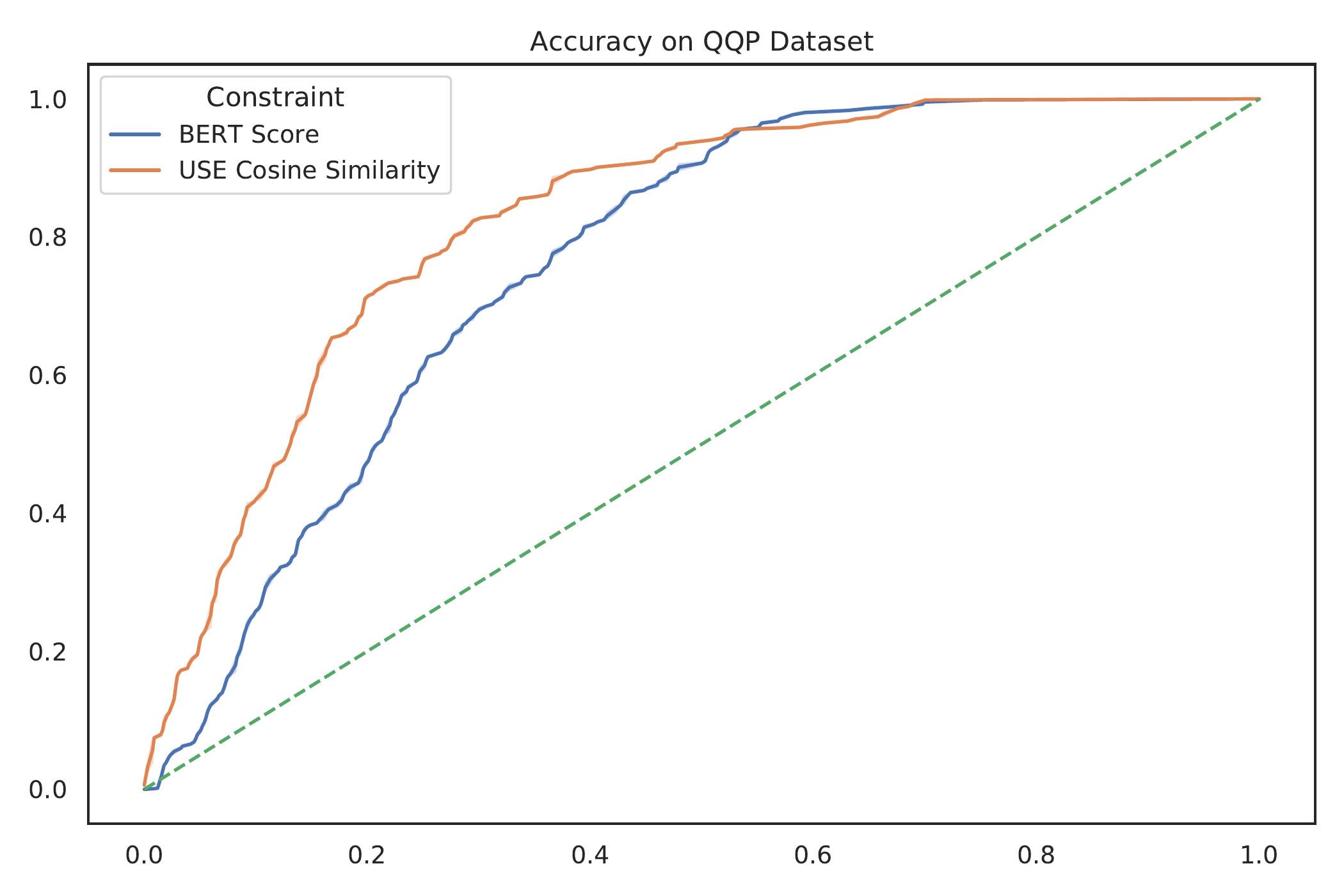}
\end{subfigure}

\begin{subfigure}{.5\textwidth}
  \includegraphics[width=\linewidth]{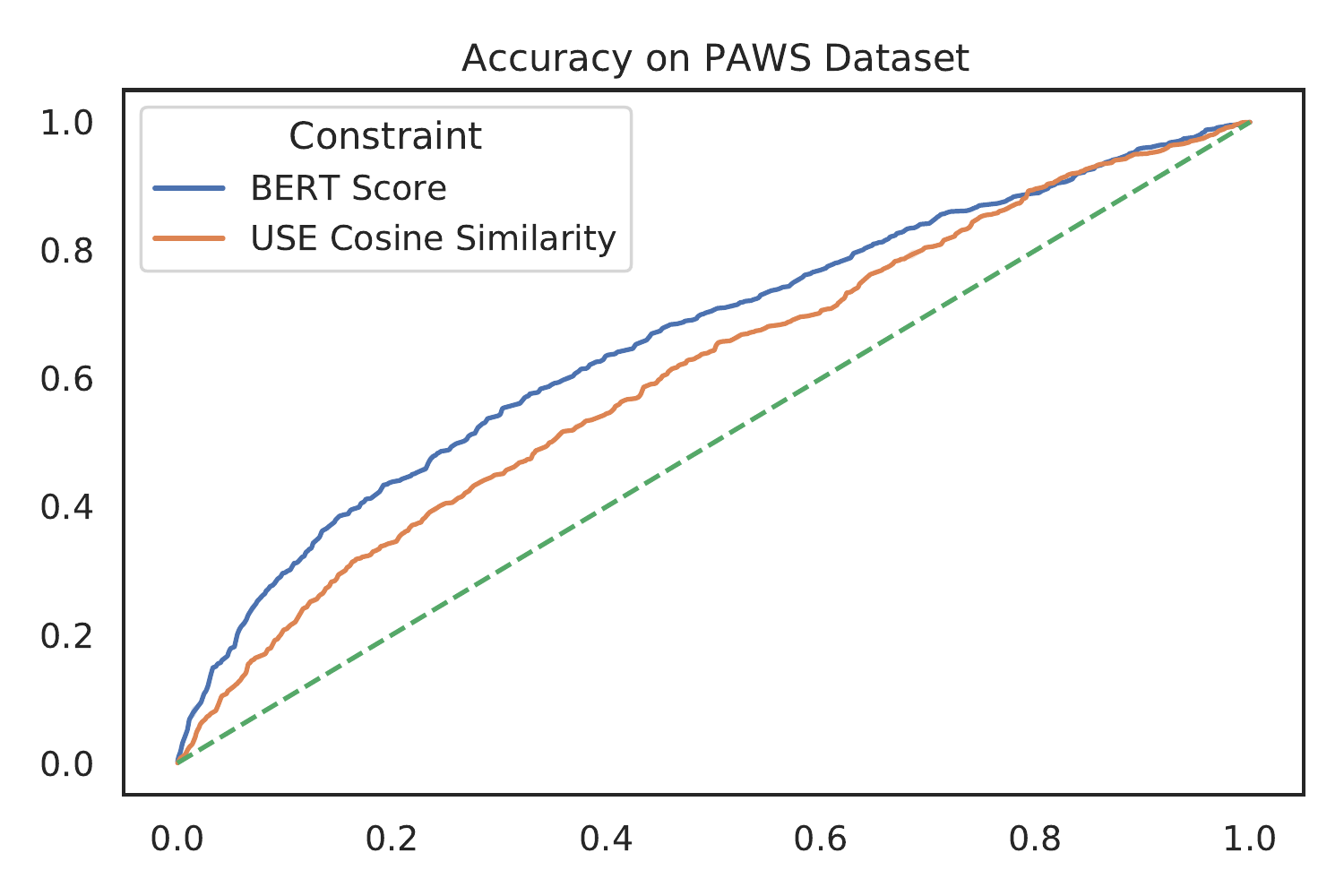}
\end{subfigure}

\caption{ROC Curves for BERTScore and the Universal Sentence Encoder (USE) on the QQP and PAWS datasets. USE outperforms BERTScore on QQP, but BERTScore is better at PAWS.}

\label{app:qqp-paws-roc}

\end{figure}

We then used these scores to plot ROC curves for each dataset; these are shown in Figure \ref{app:qqp-paws-roc}. This table shows AUC for each model and dataset. Surprisingly, USE (AUC 0.827) slightly outperforms BERTScore (AUC 0.764) on QQP; however, BERTScore (AUC 0.662) outperforms USE (AUC 0.608) on the PAWS dataset. This corroborates findings from \citet{BERTScore-Zhang2019-no} that BERTScore is superior to sentence encoding methods on datasets with high lexical overlap.

\subsubsection{Performance on Adversarial SNLI}

BERTScore exhibited higher performance than USE on PAWS, a dataset of adversarial crafted paraphrases. However, USE outperformed on QQP, a more traditional paraphrase task. To shed light on which method might perform better in an NLP attack setting, we generate a dataset that resembles potential perturbations during an NLP attack.

We set out to compare the two constraints in a scenario more similar to a typical NLP adversarial attack. To do this, we crafted a dataset of perturbations that might appear during the course of an adversarial attack. 

We crafted our dataset of adversarial perturbations starting with examples from the SNLI dataset. We chose SNLI because it is commonly used for testing NLP adversarial attack systems \cite{Survey-NLPAE-Zhang2020-cs}, and because second-order adversarial examples are particularly dangerous in the case of entailment, where a slight change in meaning can cause a shift in ground-truth output. However, this process could be emulated to test out constraint options before running an adversarial attack on any NLP dataset.

We sampled 1,000 (premise, hypothesis) from the SNLI dataset and discarded each premise. For each hypothesis, we created ten adversarial examples: one by substituting synonyms, and one by substituting antonyms, and by substituting each of $(10\%, 20\%, 30\%, 40\%, 50\%)$ of the original words. This produced a dataset with 10,000 examples. We sourced synonyms and antonyms from WordNet \cite{WordNet-Miller1995-ft}. 

BERTScore achieved a higher AUC on the two adversarial datasets, PAWS and Adversarial SNLI. This is a surprising result since BERTScore turned out to be so much less effective than USE as a constraint on adversarial examples (see Section \ref{s5:results}). We hypothesize that BERTScore is better at measuring semantic changes of 1-2 words, while USE is superior as the perturbation size grows beyond 2 words.

We can also see how across datasets, BERTScore assigns scores that are generally lower; a threshold of $\epsilon=0.8$ on USE cosine similarity may correspond to a lower threshold, for example, $\epsilon=0.5$.